\documentclass[10pt,twocolumn,letterpaper]{article}

\usepackage{cvpr}              

\usepackage[dvipsnames]{xcolor}

\definecolor{cvprblue}{rgb}{0.21,0.49,0.74}
\usepackage[pagebackref,breaklinks,colorlinks,citecolor=cvprblue]{hyperref}
\usepackage[linesnumbered, ruled]{algorithm2e}
\usepackage{tabularx}
\usepackage{multirow}
\usepackage{bm}
\usepackage{amssymb}

\usepackage{amsmath}
\usepackage{booktabs}

\usepackage{diagbox}
\usepackage{makecell}
\usepackage{colortbl}
\usepackage{wrapfig}

\newcommand\blfootnote[1]{%
\begingroup 
\renewcommand\thefootnote{}\footnote{#1}%
\addtocounter{footnote}{-1}%
\endgroup 
}

\usepackage{graphicx}
\usepackage{float}

\usepackage{arydshln}
\usepackage{multirow}
\usepackage{makecell}
\usepackage[flushleft]{threeparttable}

\newlength\savewidth

\def\paperID{14510} 
\def\confName{CVPR}
\def\confYear{2024}

\title{Small Scale Data-Free Knowledge Distillation}

\vspace{-0.5cm}
\author{\textbf{He Liu}$^{1,2 \ast}$, \textbf{Yikai Wang}$^{2 \ast}$, \textbf{Huaping Liu}$^2$, \textbf{Fuchun Sun}$^2$, \textbf{Anbang Yao}$^{3 \dag}$\vspace{0.2cm}\\
$^1$China Telecom Cloud Technology Co., Ltd.\\
$^2$BNRist Center, State Key Lab on Intelligent Technology and Systems,\\
Department of Computer Science and Technology, Tsinghua University \quad
$^3$Intel Labs China \\
{\tt\small {\{liuhe17@mails.,yikaiw@,hpliu@,fuchuns@\}tsinghua.edu.cn, anbang.yao@intel.com}}
}

\begin{document}
\maketitle

\blfootnote{$^\ast$ Equal contribution. \quad\quad $^\dag$ Corresponding author.\\
This work was done when He Liu was an intern at Intel Labs China, supervised by Anbang Yao who conceived the project.}

\begin{abstract}
Data-free knowledge distillation is able to utilize the knowledge learned by a large teacher network to augment the training of a smaller student network without accessing the original training data, avoiding privacy, security, and proprietary risks in real applications. In this line of research, existing methods typically follow an inversion-and-distillation paradigm in which a generative adversarial network on-the-fly trained with the guidance of the pre-trained teacher network is used to synthesize a large-scale sample set for knowledge distillation. In this paper, we reexamine this common data-free knowledge distillation paradigm, showing that there is considerable room to improve the overall training efficiency through a lens of ``\textbf{small-scale inverted data for knowledge distillation}". In light of three empirical observations indicating the importance of how to balance class distributions in terms of synthetic sample diversity and difficulty during both data inversion and distillation processes, we propose Small Scale Data-free Knowledge Distillation (\textbf{SSD-KD}). In formulation, SSD-KD introduces a modulating function to balance synthetic samples and a priority sampling function to select proper samples, facilitated by a dynamic replay buffer and a reinforcement learning strategy. As a result, SSD-KD can perform distillation training conditioned on an extremely small scale of synthetic samples (e.g., 10$\times$ less than the original training data scale), making the overall training efficiency one or two orders of magnitude faster than many mainstream methods while retaining superior or competitive model performance, as demonstrated on popular image classification and semantic segmentation benchmarks. The code is available at \url{https://github.com/OSVAI/SSD-KD}.
\end{abstract}    
\section{Introduction}
\label{sec:intro}

For computer vision applications on resource-constrained devices, how to learn portable neural networks yet with satisfied prediction accuracy is the key problem. Knowledge distillation (KD)~\cite{kd_hinton,kd_ba,fitnets,park2019relational,zhao2022decoupled,liu2019structured}, which leverages the information of a pre-trained large teacher network to promote the training of a smaller target student network on the same training data, has become a mainstream solution. 
Conventional KD methods assume that the original training data is always available. However, accessing the source dataset on which the teacher network was trained is usually not feasible in practice, due to its potential privacy or security or proprietary or huge-size concerns. To relax the constraint on training data, knowledge distillation under a data-free regime has recently attracted increasing attention. 

 \setlength{\abovecaptionskip}{-5pt}

\begin{figure*}
  \centering
  \begin{subfigure}{0.28\linewidth}
   \includegraphics[width=1.0\linewidth]{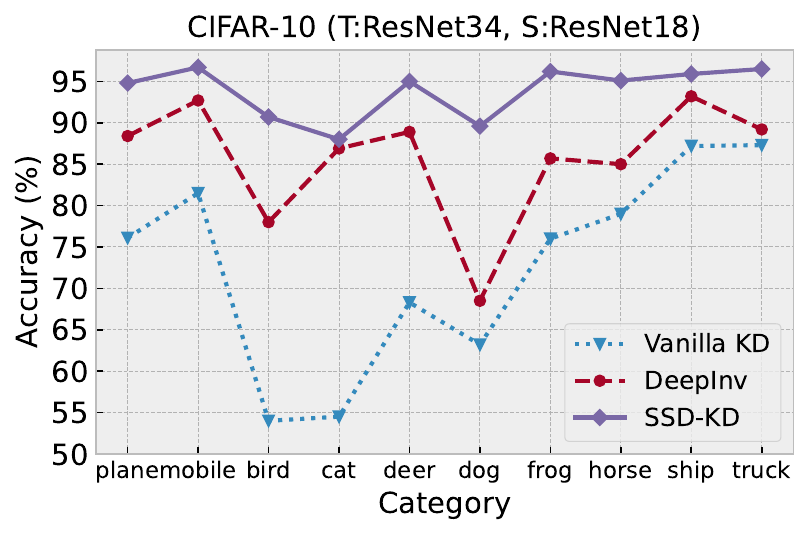}
    \label{fig:acc-res}
  \end{subfigure}
    \begin{subfigure}{0.28\linewidth}
   \includegraphics[width=1.0\linewidth]{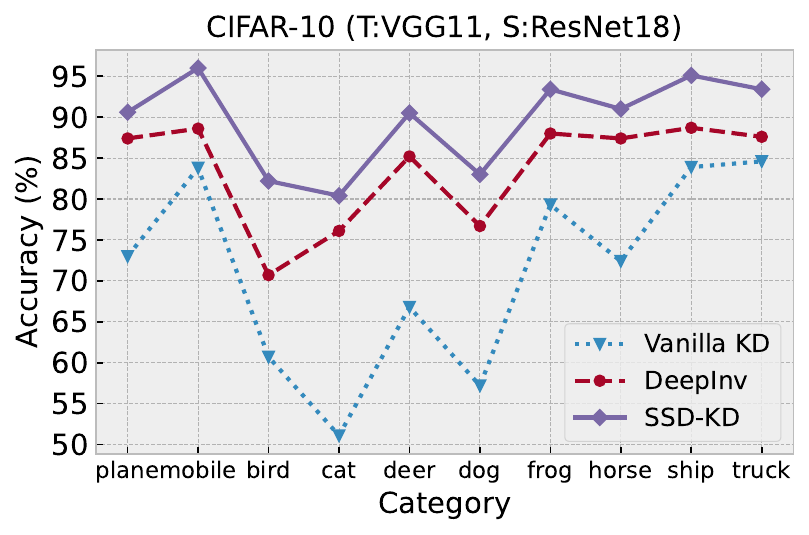}
    \label{fig:acc-vgg}
  \end{subfigure}
  \begin{subfigure}{0.28\linewidth}
   \includegraphics[width=1.0\linewidth]{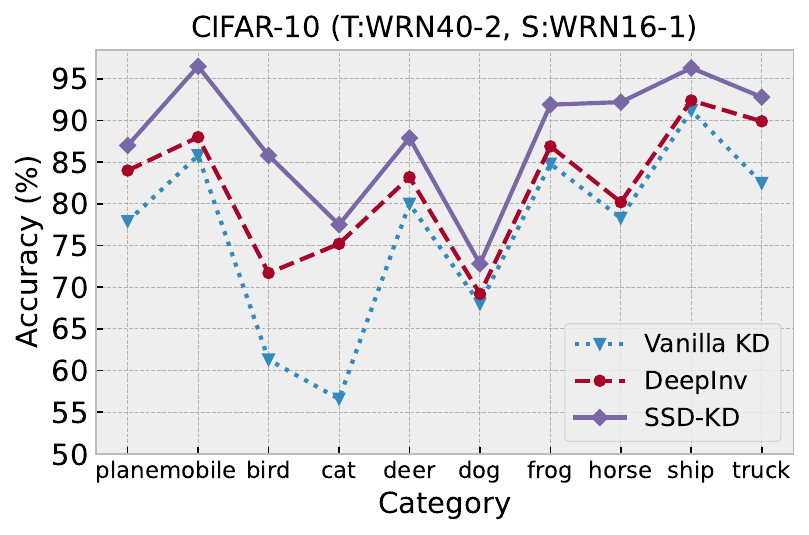}
    \label{fig:acc-wrn}
  \end{subfigure}
  \caption{Comparison of knowledge distillation (KD) using original samples vs. synthetic samples, under the same training data scale: 5000 samples (10\% of the CIFAR-10 training dataset size). In such small-scale KD regime, the student models (DeepInv~\cite{deepversionyin2020dreaming} and SSD-KD) trained on synthetic samples always show much better accuracy than the counterparts (Vanilla KD~\cite{kd_hinton}) trained on original samples.}
  \vspace{-0em}
  \label{fig:acc}
\end{figure*}

\begin{figure*}
\setlength{\abovecaptionskip}{-0.0pt}
\setlength{\belowcaptionskip}{-0.0pt}
  \centering\vspace{-0.6em}
  \begin{subfigure}{0.495\linewidth}
   \includegraphics[width=1.0\linewidth]{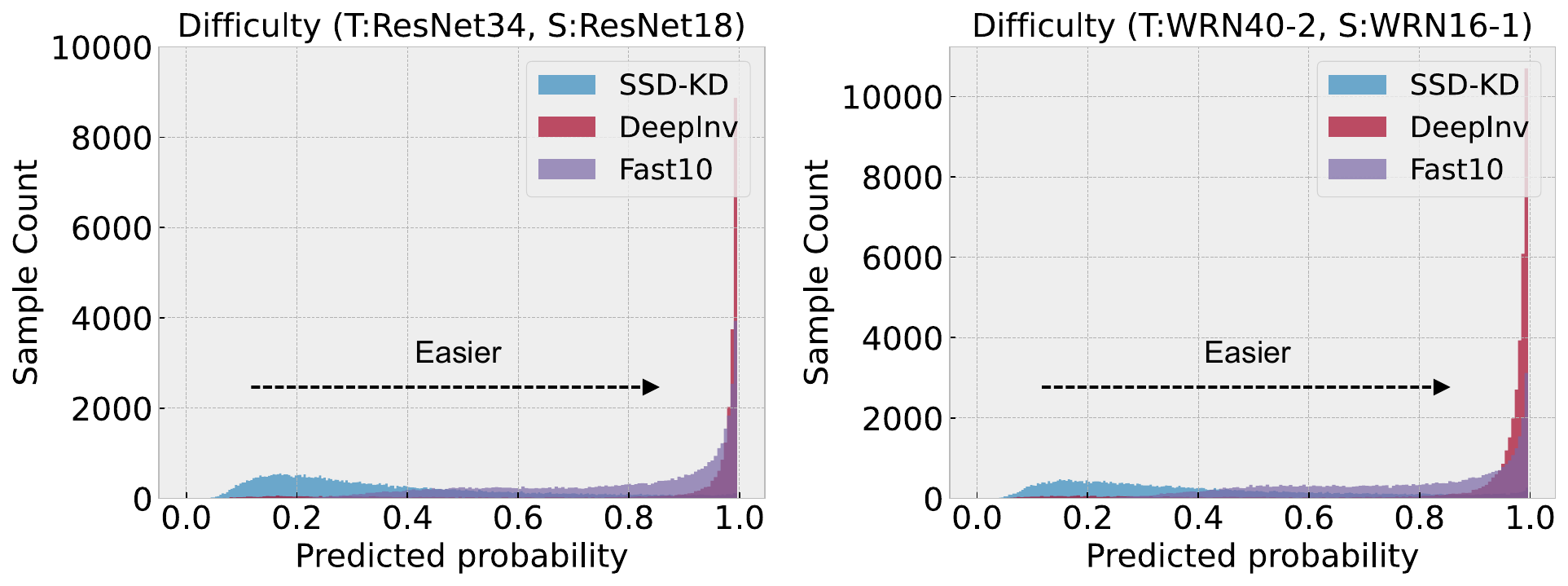}
    \caption{Difficulty distribution.}
    \label{fig:difficulty}
  \end{subfigure}
  \hfill
  \begin{subfigure}{0.495\linewidth}
   \includegraphics[width=1.0\linewidth]{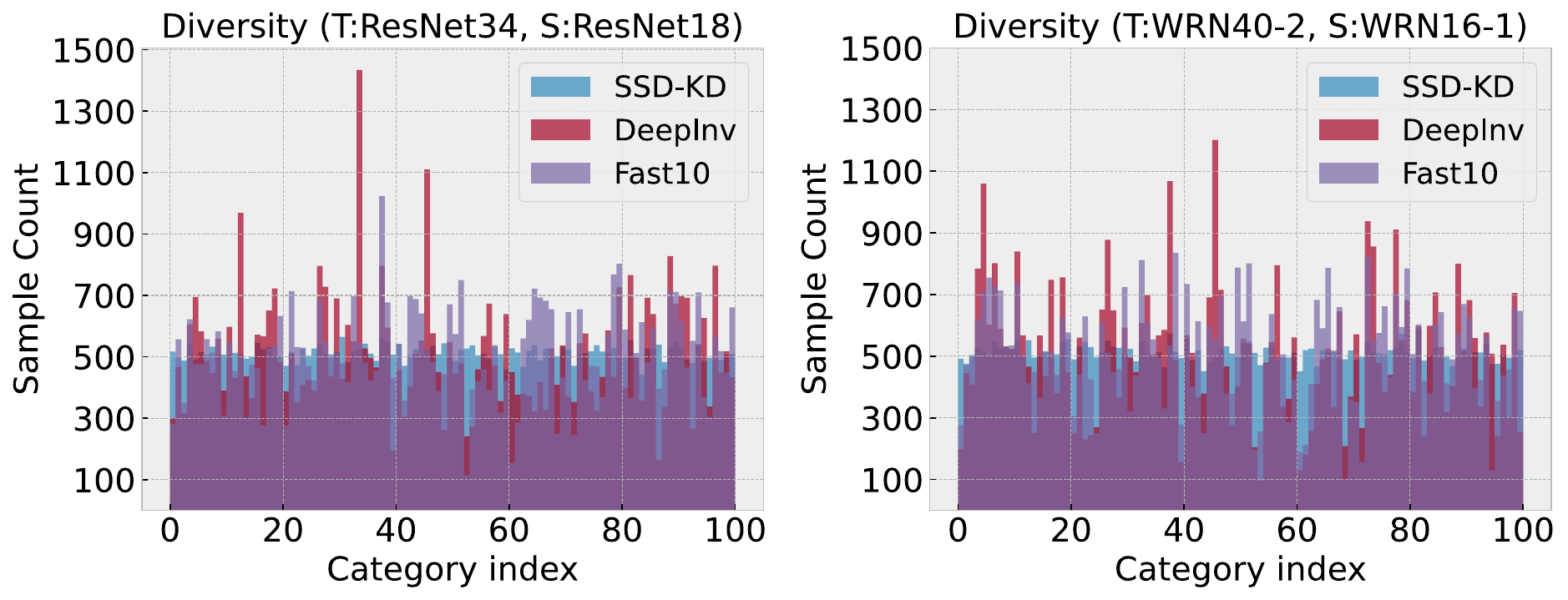}
    \caption{Diversity distribution.}
    \label{fig:diversity}
  \end{subfigure}\vspace{0.3em}
    \caption{Comparison of synthetic sample distributions collected from two top-performing adversarial D-KD methods (DeepInv~\cite{deepversionyin2020dreaming} and Fast10~\cite{fasterfang}) and our SSD-KD on CIFAR-100 dataset. Comparatively, 
    Fig.~\ref{fig:difficulty} 
    shows that our method can better balance the difficulty distribution of synthetic samples while encouraging the generator to invert more hard samples, and 
    Fig.~\ref{fig:diversity} further shows that our method can better balance the diversity distribution of synthetic samples across different categories.}
    \label{fig:sample_diversity}
    \vspace{-1.em}
\end{figure*}

The basic idea of Data-free Knowledge Distillation (D-KD)~\cite{lopes2017data,bhardwaj2019dream,zskdnayak2019zero} is to construct synthetic samples for knowledge distillation conditioned on the pre-trained teacher network, which would match the underlying distribution of the original training data. Existing top-performing D-KD methods~\cite{chen2019data,yoo2019knowledge,deepversionyin2020dreaming,dfqakdchoi2020data,cmifang2021contrastive,do2022momentum,patel2023learning} generally adopt an adversarial inversion-and-distillation paradigm. Under this paradigm, (1) during the inversion process, a generator is trained by taking the pre-trained teacher network as the discriminator; (2) during the subsequent knowledge distillation process, the on-the-fly learned generator will synthesize pseudo samples for training the student network. However, adversarial D-KD methods usually require generating a large number of synthetic samples (compared to the original training dataset size) in order to guarantee trustworthy knowledge distillation. This poses a heavy burden on training resource consumption, suppressing their use in real applications. 
In the recent work of~\cite{fasterfang}, the authors present an effective meta-learning strategy that seeks common features and reuses them as initial priors to reduce the number of iteration steps required to reach the convergence of the generator. Although faster data synthesis can be attained,~\cite{fasterfang} still needs to generate a sufficiently large number of synthetic samples to ensure effective knowledge distillation, neglecting the efficiency of the following knowledge distillation process which will become the major bottleneck to the overall training efficiency. \textit{In a nutshell, there is no research effort made to improve the overall training efficiency of D-KD via jointly considering data inversion and knowledge distillation processes, to the best of our knowledge}.

To remedy this critical gap, this paper presents the first fully efficient D-KD approach termed Small Scale Data-free Knowledge Distillation (\textbf{SSD-KD}). Our SSD-KD improves the overall training efficiency of the adversarial inversion-and-distillation paradigm from a novel ``\textbf{small data scale}'' perspective. 
In this work, ``\textbf{data scale}'' refers to the total number of inverted samples used in knowledge distillation during a training epoch. The formulation of SSD-KD is inspired by three empirical observations. On different pairs of teacher-student networks, we first observe that the student networks trained on synthetic samples tend to show much better performance than their corresponding counterparts trained on original samples when significantly reducing data scales of synthetic samples and original samples to the same (e.g., 10\% of the source training dataset size), as illustrated in Fig.~\ref{fig:acc}. Note that synthetic samples are generated with the guidance of the teacher network pre-trained on the whole source dataset, which naturally reflect different views of the original data distribution. Under a small enough data scale, this makes synthetic samples have superior capability to original samples in fitting the underlying distribution of the whole source dataset. This inspiring observation indicates that if we can construct a small-scale set of high-quality synthetic samples, a promising way toward fully efficient D-KD would be created. In principle, we believe  a high-quality small-scale synthetic dataset should have well-balanced class distributions in terms of both synthetic sample diversity and difficulty. 
However, our other two observations indicate that existing D-KD methods including both conventional and the most efficient designs do not have good capabilities to balance the aforementioned two class distributions of synthetic samples under a small data scale, as illustrated in Fig.~\ref{fig:sample_diversity}. 
Note that there already exist a few D-KD methods to enhance the diversity of synthetic samples~\cite{deepversionyin2020dreaming,dfqakdchoi2020data,cmifang2021contrastive,li2021mixmix}, but 
the diversity of synthetic samples in terms of sample difficulty is not explored yet.

Driven by the above observations and analysis, we come up with our SSD-KD which introduces two interdependent modules to significantly accelerate the overall training efficiency of the predominant adversarial inversion-and-distillation paradigm. The first module of SSD-KD relies on a novel modulating function that defines a diversity-aware term and a difficulty-aware term to jointly balance the class distributions of synthetic samples during both data synthesis and knowledge distillation processes in an explicit manner. The second module defines a novel priority sampling function facilitated by a reinforcement learning strategy that selects a small portion of proper synthetic samples from candidates stored in a dynamic replay buffer for knowledge distillation, further improving the end-to-end training efficiency. Benefiting from these two modules, SSD-KD has two appealing merits. On one side, SSD-KD can perform distillation training conditioned on an extremely small scale of synthetic samples (10$\times$ less than original training data scale), making the overall training efficiency one or two orders of magnitude faster than many mainstream D-KD methods while retaining competitive model performance. On the other side, SSD-KD attains largely improved performance in student model accuracy and maintains overall training efficiency when relaxing the data scale of synthetic samples to a relatively large number (which is still smaller than those for existing D-KD methods). We validate these merits of our method by lots of experiments on popular image classification and semantic segmentation benchmarks.
\section{Related Work}
\label{sec:formatting}

 \noindent\textbf{Data-free knowledge distillation.} D-KD is originally explored by~\cite{lopes2017data} which assumes that 
 layer-wise activation records of a well-trained teacher network pre-computed on training samples is available. The prevailing inversion-and-distillation paradigm for D-KD without reliance on the original training data or the recorded metadata is introduced in~\cite{zskdnayak2019zero}. In the inversion process, synthetic training samples are generated by leveraging the information learned by the teacher network, whose distribution is expected to fit the underlying distribution of the original training dataset. In the distillation process, the student network is trained on synthetic samples by forcing it to match the predictions of the teacher network. Based on this paradigm,~\cite{chen2019data,yoo2019knowledge} use generative adversarial networks for data inversion. Subsequent D-KD methods mostly follow this adversarial inversion-and-distillation paradigm. They try to improve the data inversion process from different aspects, such as enhancing synthetic sample discrimination with contrastive learning~\cite{cmifang2021contrastive} or an ensemble of generators~\cite{luo2020large}, combating distribution shift with momentum adversarial learning~\cite{do2022momentum} or meta learning~\cite{patel2023learning}, and promoting adversarial learning with data augmentation~\cite{li2021mixmix,yu2023data}. Our method intends to improve the overall training efficiency of the adversarial D-KD, and differs from these methods both in focus and formulation.

 \noindent\textbf{Efficient synthetic data sampling.} How to select proper synthetic samples for knowledge distillation is essential in D-KD research. Existing methods~\cite{cmifang2021contrastive,fasterfang,do2022momentum,patel2023learning} commonly rely on a memory bank to store synthetic samples, and directly update synthetic samples 
 without considering the efficiency of the following knowledge distillation process. In sharp contrast, our method introduces a reinforcement learning strategy that adaptively selects appropriate synthetic samples to update a portion of existing samples in a dynamic replay buffer by explicitly measuring their priorities in terms of jointly balancing sample diversity and difficulty, significantly improving knowledge distillation efficiency. As far as we know, our method made the first attempt to extend reinforcement learning methodology~\cite{andrychowicz2017hindsight,schaul2015prioritized,horgan2018distributed} to address data-free knowledge distillation.

\section{Method}


\subsection{Preliminaries: D-KD}
\label{section-pre}

Let ${f_t(\cdot;\theta_t)}$ be a teacher model pre-trained on the original task dataset that is no longer accessible, the goal of D-KD is to first construct a set of synthetic training samples $x$ 
via inverting the data distribution information learned by the teacher model, on which a target student model ${f_s(\cdot;\theta_s)}$ then can be trained by forcing it to mimic the teacher's function. Existing D-KD methods mostly use a generative adversarial network $g(;\theta_g)$ for producing synthetic training samples $x=g(z;\theta_g)$ from the latent noise input $z$, which is trained by taking the teacher model as the discriminator.


The optimization of D-KD contains a common distillation regularization to minimize the teacher-student function discrepancy $\mathcal{L}_{\text{KD}}({x})=D_\text{KD}(f_t(x; \theta_t)\| f_s(x; \theta_s))$\footnote{Concretely, $f_t(x; \theta_t)$ and $f_s(x; \theta_s)$ are the outputted logits (before $\mathrm{softmax})$ of the teacher model and the student model, respectively.} based on the KL-divergence, and  a task-oriented regularization $\mathcal{L}_{\text{Task}}({x})$, \emph{e.g.}, the cross-entropy loss using the teacher's predication as the ground truth. Besides, since D-KD is primarily based on the assumption that the teacher model has been optimized to be capable of capturing the source training data distribution after pre-training, recent D-KD methods~\citep{ deepversionyin2020dreaming, dfqakdchoi2020data, smith2021always} introduce an extra loss to regularize the statistics ( Batch-Normalization (BN) parameters) of the training data distribution during the data inversion process,
\begin{equation}
\mathcal{L}_{\text{BN}}({x})= \sum_l \big\|\mu_l(x)-\mathbb{E}(\mu_l)\big\|_2+\big\| \sigma_l^2(x)-\mathbb{E}(\sigma_l^2)\big\|_2,
\end{equation}
where $\mu_l(\cdot)$ and $\sigma_l(\cdot)$ denote the batch-wise mean and variance estimates of feature maps at the $l$-th layer, respectively; $\mathbb{E}(\cdot)$ over the BN statistics can be approximately substituted by running mean or variance.



\begin{figure*}
\setlength{\abovecaptionskip}{-0.0pt}
\setlength{\belowcaptionskip}{-0.0pt}
  \centering\vspace{-0.3em}
   \includegraphics[width=1.0\linewidth]{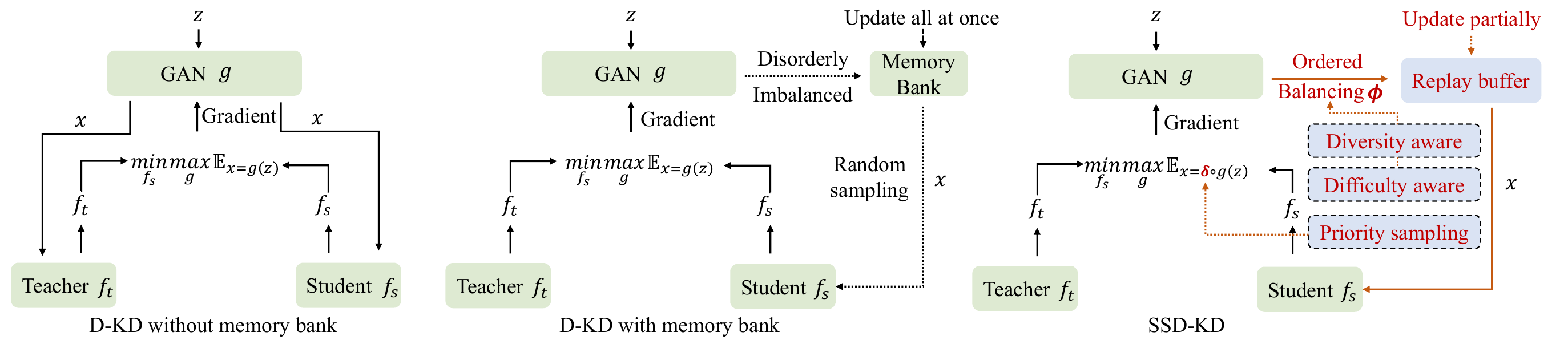}\vspace{0.3em}
    \caption{Comparison of optimization pipelines for existing adversarial D-KD methods including both conventional family~\cite{chen2019data, dfqakdchoi2020data, zskt2019, deepversionyin2020dreaming} (left) and more efficient family~\cite{cmifang2021contrastive, fasterfang} (middle), and our SSD-KD (right). Our SSD-KD formulates a reinforcement learning strategy that can flexibly seek appropriate synthetic samples to update a portion of existing samples in a dynamic replay buffer by explicitly measuring their priorities in terms of jointly balancing sample diversity and difficulty distributions. See the method section for notation definitions.} 
    \label{fig:overall-framework}
    \vspace{-1.em}
\end{figure*}

The effectiveness of D-KD heavily depends on the quality of synthetic samples which are inverted from leveraging the knowledge of the pre-trained teacher model. The prevailing adversarial D-KD paradigm consists of two processes, namely data inversion and knowledge distillation. From both perspectives of efficiency and efficacy, on the one hand, the data inversion process largely affects the optimization performance of the student model; on the other hand, the training time cost of knowledge distillation emerges as a significant constraint to the overall training efficiency of D-KD.

\subsection{Our Design: SSD-KD}

Regarding our SSD-KD, we focuses on improving the adversarial D-KD paradigm through a lens of ``\textbf{small-scale inverted data for knowledge distillation}". SSD-KD lays emphasis on instructing the data inversion process with the feedback of both the pre-trained teacher and the knowledge distillation process, significantly accelerating the overall training efficiency. 
Following the notations in the prevision subsection, the optimization goal of our SSD-KD is defined as 
\begin{align}
\resizebox{0.9\hsize}{!}{ $
\min\limits_{f_s} \max\limits_{g}  \mathbb{E}_{x = \delta \circ g\left(z\right)}\big(\mathcal{L}_{\text{BN}}({x})+\mathcal{L}_{\text{KD}}({x})+\phi(x)\mathcal{L}_{\text{Task}}({x})\big),
\label{eq:loss}
$}
\end{align}
where,
\begin{itemize}
\item  We adopt a diversity-aware modulating function $\phi(x)$ that allocates each synthetic sample with a different priority regarding its predicted category by the teacher model, as presented in Sec.~\ref{sec:distribution-aware}.
\item Under the constraint of BN estimates, with $\phi(x)$, we encourage the generator to explore as tough synthetic samples (\emph{w.r.t.} the teacher model) as possible, as introduced in Sec.~\ref{sec:distribution-aware}.
\item Instead of applying a random sampling strategy to select samples for knowledge distillation, we adopt a re-weighting strategy to control the sampling process. We abuse notation slightly with $\circ$ to represent applying the strategy based on priority sampling function $\delta$, with more details in Sec.~\ref{sec:sorting}.
\item Each synthetic sample is not only prioritized by its modulating function $\phi(x)$ but also is reweighted at the sampling stage that reuses the same intermediate values as $\phi(x)$.
\end{itemize}





Although the D-KD pipeline allows training samples to be synthesized and served for training the student model on the same task. However, there is a large extent of data redundancy that hinders the training efficiency of D-KD methods. 
In the following sections, we detail our SSD-KD, a fully efficient D-KD method that is capable of using an extremely small scale of synthetic data yet achieving competitive performance compared to existing D-KD methods.

The pipeline of SSD-KD is summarized in Alg.~\ref{alg:cmi} and the comparison of optimization pipelines for existing adversarial D-KD methods including both conventional family~\cite{chen2019data, dfqakdchoi2020data, zskt2019, deepversionyin2020dreaming} and more efficient family~\cite{cmifang2021contrastive, fasterfang}, and our SSD-KD is shown in Fig. \ref{fig:overall-framework}.

\subsection{Data Inversion with Distribution Balancing}
\label{sec:distribution-aware}

We provide Fig.~\ref{fig:sample_diversity} to demonstrate the data redundancy of D-KD that results from a large imbalance of the synthetic data. The left two sub-figures of Fig.~\ref{fig:sample_diversity} depict the distribution of categories predicted by the teacher model,  indicating a significant imbalance in data categories. The right two sub-figures of Fig.~\ref{fig:sample_diversity} show sample accounts of different bars that correspond to different prediction difficulties (the difficulty is measured by the predicted probability by the teacher model). For D-KD, it indicates that generating samples with only the instruction of teacher-student discrepancy results in a sharp distribution over the sample difficulty and tends to obtain easily predicted data samples. We argue that for the D-KD task where data samples are all synthesized, the data-generating process needs to consider both the teacher-student discrepancy and the pre-trained knowledge of the teacher itself, by which our SSD-KD proposes diversity-aware and difficulty-aware data synthesis, as detailed below.

\textbf{Diversity-aware balancing.} We first propose to address the issue of imbalanced sample difficulty in the data inversion process.  Specifically, we maintain a replay buffer $\mathcal{B}$ that stores a constant amount (denoted as $|\mathcal{B}|$) of synthetic data samples. For each data sample $x$ in $\mathcal{B}$, we penalize its total amount of samples that share the same predicted category (by the teacher model) with $x$. To realize this, we adopt a diversity-aware balancing term that encourages the generator to synthesize samples with infrequent categories, which will be shown in Eq.~(\ref{eq:phi}).

\textbf{Difficulty-aware balancing.}
Drawing inspiration from the field of object detection that utilizes focal loss for largely imbalanced samples~\cite{focallosslin2017focal, OHEMshrivastava2016training}, for each sample $x$, we further introduce a difficulty-aware balancing term on the predicted probability $p_T(x)$. Here, difficult synthetic samples are considered as those with low-confidence predictions by the teacher model, which are encouraged by the difficulty-aware balancing term as will be given in Eq.~(\ref{eq:phi}).

In summary, we introduce a modulating function $\phi(x)$ to adjust the optimization of the generator based on the prediction feedback from the pre-trained teacher model. $\phi(x)$ is expected to balance the category distribution and dynamically distinguish between easy and hard synthetic samples, by which easy samples no longer overwhelm the distillation process. Formally, for a synthetic data sample $x\in\mathcal{B}$, its modulating function $\phi(x)$ is computed by
\begin{equation}
\resizebox{0.9\hsize}{!}{ $
    \phi (x) = \underbrace {\Big(1 - \frac{1}{|\mathcal{B}|}\sum_{x'\in\mathcal{B}}\mathbb{I}_{c_T(x')=c_T(x)}\Big)}_{\text{diversity-aware balancing}} \underbrace{{\Big(1 - p_T(x)\Big)}^\gamma }_{\text{difficulty-aware balancing}},$
    }
\label{eq:phi}
\end{equation}
where $c_T(x)$ and  $p_T(x)$ refer to the predicted category index and probability (or confidence) by the pre-trained teacher model, respectively; $\mathbb{I}_{c_T(x')=c_T(x)}$ denotes an indicator function that equals to $1$ if the predicted category of $x'$ is the same with that of $x$, while otherwise $0$; $\gamma$ is a hyper-parameter.

We highlight two properties of the modulating function $\phi (x)$. Firstly, for the data sample $x$ with a high prediction certainty \emph{w.r.t.} the teacher model (\emph{i.e.}, considered as an easy sample), $\phi (x)$ approaches to a low value and thus results in low impact on the task-oriented loss $\mathcal{L}_{\text{Task}}({x})$ in Eq.~(\ref{eq:loss}). Secondly, when the category distribution of synthetic data samples in $\mathcal{B}$ is largely imbalanced as predicted by the teacher network, the sample $x$ of which the category shares with more samples in  $\mathcal{B}$ is penalized and thus the corresponding $\mathcal{L}_{\text{Task}}({x})$ is weakened by $\phi (x)$.

Although Eq.~(\ref{eq:phi}) implies that the value of the modulating function $\phi (x)$ is partially determined by the current replay buffer $\mathcal{B}$, note that $\mathcal{B}$ changes dynamically and is also affected by $\phi (x)$. This is due to that the term $\phi (x)\mathcal{L}_{\text{Task}}({x})$ in Eq.~(\ref{eq:loss}) directly optimizes the generator that synthesizes data samples to compose $\mathcal{B}$. In this sense, the balancing in category diversity is maintained during training given the mutual effects of $\mathcal{B}$ and $\phi (x)$. We find it particularly important in the data inversion process for balancing categories of data samples.

With both balancing terms, as shown in Fig.~\ref{fig:sample_diversity}, our method (SSD-KD) generates mild sample distributions in terms of both sample category and difficulty.

\begin{algorithm}[t!]
   \setlength{\abovecaptionskip}{-8pt}
   \setlength{\belowcaptionskip}{-10pt}
   \KwIn{Number of training epochs $E$; number of iterations for data inversion $T_{g}$; number of iterations for distillation ${T_{kd}}$; pre-trained teacher model ${f_t}(\cdot;{\theta_t})$; student network ${f_s}(\cdot;{\theta_s})$}
   \KwOut{The optimized student model ${f_s}(\cdot;{\theta_s})$}
   {}
   Initialize replay buffer $\mathcal{B} \leftarrow \emptyset$ and ${f_s}(\cdot;{\theta_s})$
  
   \For{$e = 1$ to $E$}
   {
      (1) Data Inversion with Distribution Balancing:
     
      Initialize a generative network ${g}(\cdot;{\theta_g})$
      
      $z \leftarrow \mathcal{N}(0, 1)$
      
      \For{$i = 1$ to $T_g$}
      {
         $x \leftarrow g(z; \theta_g)$
         
         Compute the overall loss by Eq.~(\ref{eq:loss})
         
         Update $z$ and $\theta_g$ 
      }

     (2) Distillation with Priority Sampling:
      
      Compute the importance-sampling weight $w_{i-1}(x)$ 
      
      Compute the sample priority $\delta_{i}(x)$
      
      $\mathcal{B} \leftarrow \mathcal{B} \cup \{x, \delta_{i}(x)\}$
      
      Remove old samples to keep $|\mathcal{B}|$ a constant
    
     \For{$j = 1$ to $T_{kd}$}
      {
        Sample a small-scale mini-batch $M$ from $\mathcal{B}$
        
        Compute the distillation loss $\mathcal{L}_{\text{KD}}({x})$ in Eq.~(\ref{eq:loss}), where $x \in M$
        
        Compute the importance-sampling weight $w_{i}(x)$ 
      
      Compute the sample priority $\delta_{i+1}(x)$

            Update old samples $\{x, \delta_{i}(x)\}$ to $\{x, \delta_{i+1}(x)\}$ in $\mathcal{B}$
        
        Update $\theta_s$  
      }
   }

   Return ${f_s}(\cdot;{\theta_s})$

   \caption{SSD-KD}
   \label{alg:cmi}
\end{algorithm}

\subsection{Distillation with Priority Sampling}
\label{sec:sorting}

The original prioritized experience replay method~\cite{schaul2015prioritized} reuses important transitions more frequently and learns more efficiently. Differently, rather than obtaining the reward from the environment, our prioritized sampling method is designed to fit in data-free knowledge distillation and get feedback from the framework itself. In other words, the prioritized sampling method performs the opposite role of the previous data-free knowledge distillation methods: it focuses on training a sparse set of highly prioritized samples instead of uniform sampling to speed up training.

By Eq.~(\ref{eq:phi}), we sample the synthetic data $x$ from the current replay buffer $\mathcal{B}$. Instead of uniformly sampling $x$, we propose to modulate the sampling probability by a sampling strategy termed Priority Sampling (PS). The basic function of PS is to measure the importance of each sample $x$ in $\mathcal{B}$, by which we introduce a priority sampling function $\delta_{i}(x)$,
\begin{equation}
    \delta_{i}(x) = w_{i-1}(x) KL(f_t(x;\theta_t)||f_s(x;\theta_s)),
\end{equation}
where as mentioned in Sec.~\ref{section-pre}, $KL$ denotes the KL-divergence between the $\mathrm{softmax}$ outputs of logits $f_t(x;\theta_t)$ and $f_s(x;\theta_s)$; $\theta_t$, $\theta_s$ depend on the training step $i$; $w_{i}(x)$ is the calibration term \cite{schaul2015prioritized} for normalizing samples in $\mathcal{B}$, as will be formalized in Eq.~(\ref{eq_wi}), especially, when $i=0$, $w_{-1}(x)=1$.

\begin{table*}
\setlength{\abovecaptionskip}{0.1in}
  \centering\vspace{-0.0em}
\centering
\caption{Performance comparison of Fast2 [11] (the current most efficient D-KD method) and our SSD-KD, in terms of top-1 classification accuracy (\%) and overall training time cost (hours). Our SSD-KD performs with a very small training data scale: 5000 synthetic samples, i.e., 10\% of the original training dataset size. All image classification results in this and the other tables are averaged over three independent runs, and the methods ``Teacher'' and ``Student'' are performed on the whole original training dataset.}
\label{tab:few-samples}
\resizebox{0.7\linewidth}{!}{%
\begin{tabular}{ccccccc}
\toprule[1.2pt]
\multirow{2}{*}{Dataset} & \multirow{2}{*}{Method} & ResNet-34 & VGG-11 & WRN40-2 & WRN40-2 & WRN40-2 \\
                           &           & ResNet-18    & ResNet-18    & WRN16-1     & WRN40-1      & WRN16-2     \\ \hline
\multirow{5}{*}{CIFAR-10}  & Teacher   & 95.70         & 92.25         & 94.87         & 94.87          & 94.87         \\
                           & Student   & 95.20         & 92.20         & 91.12         & 93.94          & 93.95         \\
                           & Fast2~\cite{fasterfang}      & 92.62(3.92h) & 84.67(3.02h) & 88.36(1.46h) & 89.56(2.37h)  & 89.68(1.70h) \\
                           & \textbf{SSD-KD} & \textbf{92.92(1.11h)} & \textbf{89.56(0.77h)} & \textbf{88.07(0.77h)} & \textbf{90.12(0.96h)}  & \textbf{91.84(0.78h)} \\ \hline
\multirow{5}{*}{CIFAR-100} & Teacher   & 78.05         & 71.32         & 75.83         & 75.83          & 75.83         \\
                           & Student   & 77.10         & 77.10         & 65.31         & 72.19          & 73.56         \\
                           & Fast2~\cite{fasterfang}      & 69.76(4.02h) & 62.83(2.99h) & 41.77(1.44h) & 53.15(2.26h)  & 57.08(1.72h) \\
                           & \textbf{SSD-KD} & \textbf{73.05(1.12h)} & \textbf{66.66(0.78h)} & \textbf{51.96(0.77h)} & \textbf{59.60(0.97h)} & \textbf{61.44(0.78h)} \\ \hline
\toprule[1.2pt]
\vspace{-0.25in}
\end{tabular}
}
\end{table*}

The training of knowledge distillation with random updates relies on those updates corresponding to the same distribution as its expectation. Prioritized sampling data introduces bias since it might change the data distribution, and affect the solution that the estimates will converge to. Thus we correct the bias by introducing an importance-sampling (IS) weight $w_i(x)$ for the data sample $x$:
\begin{equation}
\label{eq_wi}
w_{i}(x)=(N \cdot P_{i}(x))^{-\beta},
\end{equation}
where $\beta$ is a hyper-parameter; $P_{i}(x)$ is the probability of sampling transition  defined by

\begin{equation}
\label{pi}
P_{i}(x)=\frac{\big(\left|\delta_{i}(x)\right|+\epsilon\big)^\alpha}{\sum_{x'\in\mathcal{B}}\big(\left|\delta_{i}(x')\right|+\epsilon\big)^\alpha},
\end{equation}
where $\epsilon$ is a small positive constant which prevents the edge-case of transitions not being selected once their priority is zero.

The priority sampling function $\delta(x)$ has two noteworthy properties. Firstly, as the delta value increases, $\delta(x)$ reflects a greater information discrepancy between the teacher and student models for the synthetic samples in the current $\mathcal{B}$. The student model should therefore be optimized from samples with greater information discrepancy, as this facilitates the faster acquisition of the teacher model. Secondly, $\delta(x)$ dynamically changes with each update iteration of the student and generative models. Consequently, when the student model acquires the teacher model's capabilities on certain samples, it continues to learn from samples with larger differences relative to the teacher model based on the new sample distribution. This further enhances the performance of the student model.
\section{Experiment}
\label{experiment}


\begin{table*}
\setlength{\abovecaptionskip}{0.1in}
  \centering\vspace{-0.0em}
\centering
\caption{Performance comparison of existing top-performing D-KD methods and our SSD-KD, in terms of top-1 classification accuracy (\%) and overall training time cost (hours). For our SSD-KD, we relax the data scale of synthetic samples to be similar to that for Fast5.
}
\label{tab:overall-accuracy}
\resizebox{0.85\linewidth}{!}{%
\begin{tabular}{ccccccc}
\toprule[1.pt]
\multirow{2}{*}{Dataset} & \multirow{2}{*}{Method} & ResNet-34 & VGG-11 & WRN40-2 & WRN40-2 & WRN40-2 \\
                            &           & ResNet-18      & ResNet-18      & WRN16-1      & WRN40-1       & WRN16-2       \\ \hline
\multirow{11}{*}{CIFAR-10}  & Teacher   & 95.70         & 92.25         & 94.87         & 94.87          & 94.87         \\
                           & Student   & 95.20         & 92.20         & 91.12         & 93.94          & 93.95         \\
                            & DeepInv~\cite{deepversionyin2020dreaming}   & 93.26(25.62h)  & 90.36(13.19h)  & 83.04(10.62h) & 86.85(13.84h)  & 89.72(11.67h)  \\
                            & CMI~\cite{cmifang2021contrastive}      & 94.84(15.49h)  & 91.13(10.99h)  & 90.01(9.87h)  & 92.78(11.57h)  & 92.52(10.58h)  \\
                            & DAFL~\cite{chen2019data}      & 92.22(303.50h) & 81.10(209.78h) & 65.71(96.57h) & 81.33(157.44h) & 81.55(118.52h) \\
                            & DFQ~\cite{dfqakdchoi2020data}       & 94.61(152.16h) & 90.84(104.93h) & 86.14(48.40h) & 91.69(78.81h)  & 92.01(59.40h)  \\
                            & ZSKT~\cite{zskt2019}      & 93.32(304.11h) & 89.46(209.65h) & 83.74(96.75h) & 86.07(157.51h) & 89.66(118.82h) \\
                            & Fast5~\cite{fasterfang}     & 93.63(4.34h)   & 89.94(3.15h)   & 88.90(1.61h)  & 92.04(2.44h)   & 91.96(1.89h)   \\
                            & Fast10~\cite{fasterfang}    & 94.05(4.72h)   & 90.53(3.22h)   & 89.29(1.72h)  & 92.51(2.48h)   & 92.45(2.05h)   \\
                            & \textbf{SSD-KD} & \textbf{94.26(4.20h)}   & \textbf{90.67(2.95h)}   & \textbf{89.96(1.48h)}  & \textbf{93.23(2.30h)}   & \textbf{93.11(1.78h)}   \\ \hline
\multirow{11}{*}{CIFAR-100} & Teacher   & 78.05         & 71.32         & 75.83         & 75.83          & 75.83         \\
                           & Student   & 77.10         & 77.10         & 65.31         & 72.19          & 73.56         \\
                            & DeepInv~\cite{deepversionyin2020dreaming}   & 61.32(25.88h)  & 54.13(13.13h)  & 53.77(10.62h) & 61.33(13.86h)  & 61.34(11.68h)  \\
                            & CMI~\cite{cmifang2021contrastive}      & 77.04(15.42h)  & 70.56(11.28h)  & 57.91(10.01h) & 68.88(11.57h)  & 68.75(10.53h)  \\
                            & DAFL~\cite{chen2019data}      & 74.47(303.54h) & 54.16(209.82h) & 20.88(96.62h) & 42.83(157.49h) & 43.70(118.67h) \\
                            & DFQ~\cite{dfqakdchoi2020data}       & 77.01(152.18h) & 66.21(105.14h) & 51.27(48.47h) & 54.43(78.83h)  & 64.79(59.48h)  \\
                            & ZSKT~\cite{zskt2019}      & 67.74(304.13h) & 54.31(209.63h) & 36.66(96.87h) & 53.60(157.68h) & 54.59(118.84h) \\
                            & Fast5~\cite{fasterfang}     & 72.82(4.35h)   & 65.28(3.06h)   & 52.90(1.56h)  & 61.80(2.35h)   & 63.83(1.86h)   \\
                            & Fast10~\cite{fasterfang}    & 74.34(4.50h)   & 67.44(3.12h)   & 54.02(1.65h)  & 63.91(2.42h)   & 65.12(1.95h)   \\
                            & \textbf{SSD-KD} & \textbf{75.16(4.22h)}   & \textbf{68.77(2.94h)}   & \textbf{55.61(1.52h)}  & \textbf{64.57(2.27h)}   & \textbf{65.28(1.78h)}   \\ \hline
\toprule[1.pt]
\vspace{-0.25in}
\end{tabular}
}
\end{table*}

\subsection{Experimental Details}

We conduct comprehensive experiments on image classification and semantic segmentation tasks to evaluate the effectiveness and study the design of our SSD-KD.

\textbf{Datasets.} For image classification experiments, we 
use CIFAR-10 and CIFAR-100 \cite{cifar100}, two most popular datasets for D-KD research. For semantic segmentation experiments, we use the NYUv2 dataset \cite{nyuv2}. On each dataset, the baseline models are trained with its standard data split.


\textbf{Training Setups.} We follow the settings of \cite{fasterfang} for basic experiments and comparisons. Regarding experiments on the CIFAR-10 and CIFAR-100 datasets, we use 5 different teacher-student model pairs having either the same type or different type network architectures (see Table \ref{tab:few-samples}, \ref{tab:overall-accuracy}). Regarding experiments
on the NYUv2 dataset, we use two Deeplabv3 models \cite{chen2019rethinking} as a teacher-student model pair (see Table \ref{tab:segmentation-table}). Unless otherwise stated, we always adopt the same basic settings as in \cite{fasterfang} for experiments, including the number of training epochs, the optimizer, the weight decay, etc. 

\textbf{Evaluation Metrics.} Besides comparing the student model accuracy (top-1 accuracy for image classification and mean Intersection over Union (IoU) for semantic segmentation), we also compare the overall training time cost of existing mainstream D-KD methods \cite{deepversionyin2020dreaming,cmifang2021contrastive,chen2019data,dfqakdchoi2020data,zskt2019,fasterfang} and our SSD-KD. For each teacher-student model pair, we meticulously record the total time cost (hours) for each run of the end-to-end training by all methods. In order to guarantee a fair comparison, all training speed assessments are performed on 1 NVIDIA V100 GPU using 12 cores of Intel Xeon Gold 6240R CPU. Our experiments are implemented with the PyTorch~\cite{neuripsPaszkeGMLBCKLGA19} library. For each teacher-student model pair, the experiment is conducted with three independent runs for all methods, and in comparison we report the averaged results, unless  otherwise stated.

\subsection{Experimental Results}
\textbf{Results on image classification task.}  As the main focus of our work is to improve the overall training efficiency of the adversarial data-free knowledge distillation paradigm, we first compare the proposed SSD-KD with the current most efficient D-KD method Fast2 \cite{fasterfang} on the CIFAR-10 and CIFAR-100 datasets. From the results shown in Table \ref{tab:few-samples}, we can observe: (1) on the CIFAR-10 dataset, our SSD-KD (using a very small data scale: 5000 synthetic samples, i.e., 10\% of the original training dataset size) shows large improvements in training efficiency to Fast2 (at least 1.90$\times$ and at most 3.92$\times$ training speedup, on 5 teacher-student model pairs), and gets better student models on 4 out of 5 teacher-student model pairs; (2) on the more challenging CIFAR-100 dataset, our SSD-KD shows a very similar training speedup trend against Fast2 as on the CIFAR-10 dataset, but gets significantly better student models on all 5 teacher-student model pairs (at least 3.29\% and at most 10.19\% absolute top-1 accuracy gain to Fast2). The superior performance of SSD-KD against Fast2 validates the efficacy of our small-scale data inversion and sampling mechanism which can flexibly balance class distributions in terms of synthetic sample diversity and difficulty during both data inversion and distillation processes.

Next, we compare the proposed SSD-KD with existing top-performing D-KD methods including DeepInv~\cite{deepversionyin2020dreaming}, CMI~\cite{cmifang2021contrastive}, DAFL~\cite{chen2019data}, DFQ~\cite{dfqakdchoi2020data}, ZSKT~\cite{zskt2019}, Fast5 and Fast10~\cite{fasterfang}. In order to get improved student model performance, we relax the data scale of synthetic samples in our SSD-KD to be similar to that for Fast5. Experimental results are summarized in Table \ref{tab:overall-accuracy}. Compared to these mainstream D-KD methods, our SSD-KD gets very competitive performance in student model accuracy, and significantly better performance in overall training efficiency, on all 5 teacher-student model pairs.

\begin{table}
\setlength{\abovecaptionskip}{0.1in}
  \centering\vspace{-0.0em}
\centering
    \captionof{table}{Performance comparison on the NYUv2 dataset. The teacher model is pre-trained on the ImageNet dataset and fine-tuned on the NYUv2 dataset, and the student model is trained from scratch. The results of reference methods are collected from~\cite{fasterfang}.} 
    \vspace{-0.1em}
    \label{tab:segmentation-table}
    \resizebox{0.95\linewidth}{!}{%
    \begin{tabular}{ccc}
    \toprule[1pt]
    Method  & Training Data Scale    & mIoU (\%)  \\ \hline
    Teacher: Deeplabv3-ResNet50 & 1,449 NYUv2   & 0.519 \\
    Student: Deeplabv3-Mobilenetv2 & 1,449 NYUv2   & 0.375 \\
    KD~\cite{kd_hinton}      & 1,449 NYUv2   & 0.380 \\
    DFND~\cite{chen2021learning}    & 14M (ImageNet)  & 0.378 \\ \hline
    DFAD~\cite{dfadfang2019data}    & 960K (synthetic)      & 0.364 \\
    DAFL~\cite{chen2019data}    & 960K (synthetic)      & 0.105 \\
    Fast10~\cite{fasterfang}  & 17K (synthetic)  & 0.366 \\
    \textbf{SSD-KD}  & \textbf{16K} (synthetic) & \textbf{0.384} \\ \hline
    \toprule[1pt]
    \end{tabular}}
    \vspace{-1.5em}
\end{table}

In light of the results in Table \ref{tab:few-samples}, \ref{tab:overall-accuracy}, it can be seen that our SSD-KD makes the overall training efficiency one or two orders of magnitude faster than most D-KD methods while retaining competitive student model performance.



\textbf{Results on semantic segmentation task.} To further validate the generalization ability of our method, we compare the performance of SSD-KD with existing D-KD methods on the NYUv2 dataset.
Table \ref{tab:segmentation-table} summarizes the results. 
We can see that our SSD-KD not only achieves state-of-the-art performance in terms of model accuracy, but also is significantly more efficient than other D-KD methods in terms of the training data scale. The overall training time cost for our SSD-KD and Fast10 is 8.9 hours and 9.5 hours, respectively. Besides, the student model trained by SSD-KD even outperforms the baseline model trained on the original training data, as NYUv2 is a small-scale dataset.

\subsection{Ablation Studies}
\vspace{-0.em}

\begin{figure*}
\vspace{-0.3 em}
\setlength{\abovecaptionskip}{0.0in}
    \centering
    \begin{tabular}{cccc}
    \includegraphics[width=0.23\linewidth]{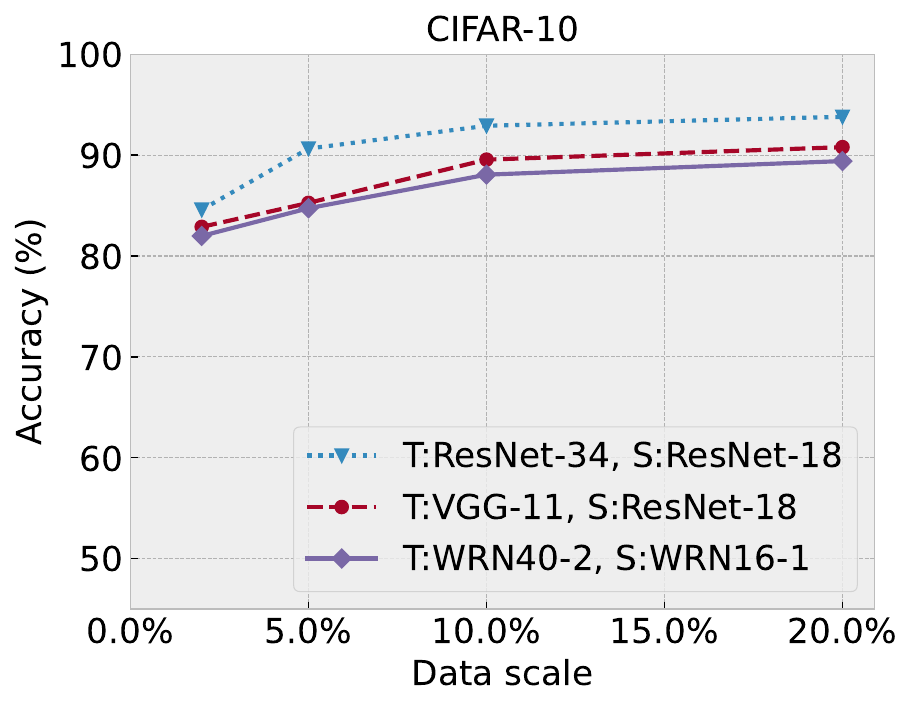}&%
    \includegraphics[width=0.23\linewidth]{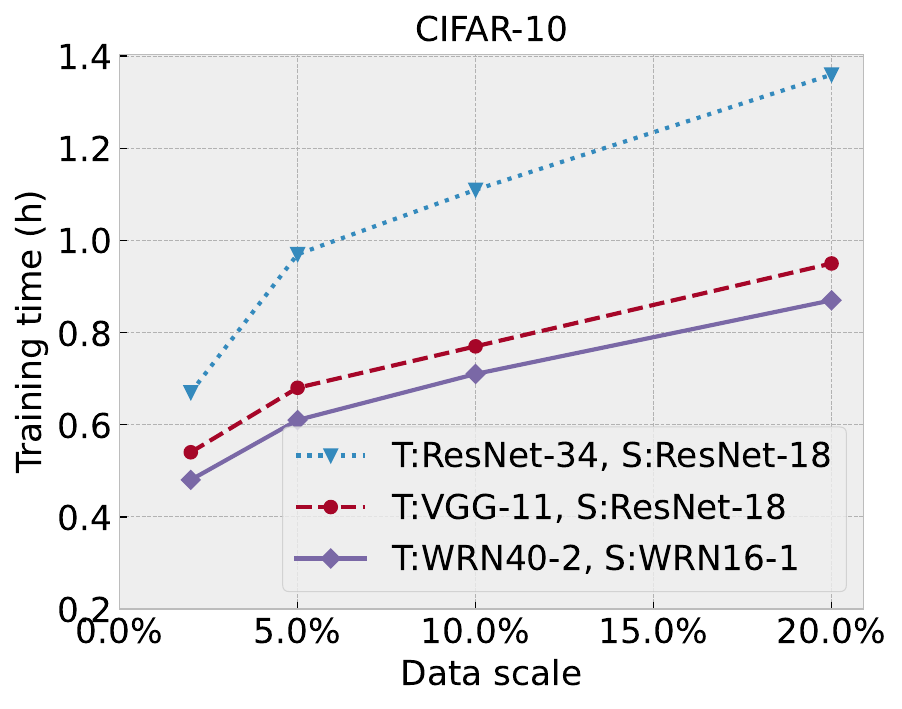}&%
    \includegraphics[width=0.23\linewidth]{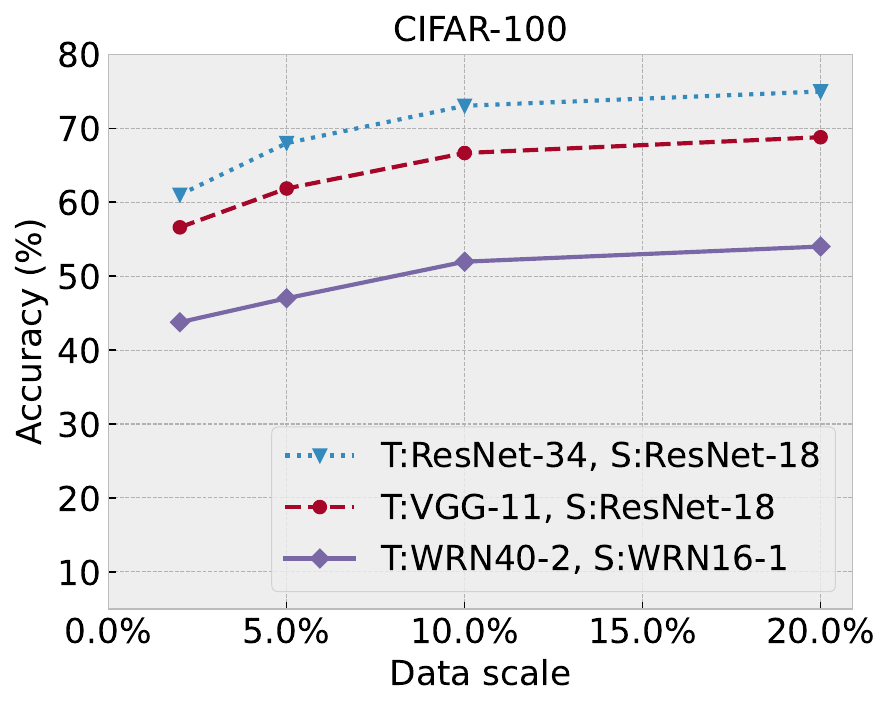}&%
    \includegraphics[width=0.23\linewidth]{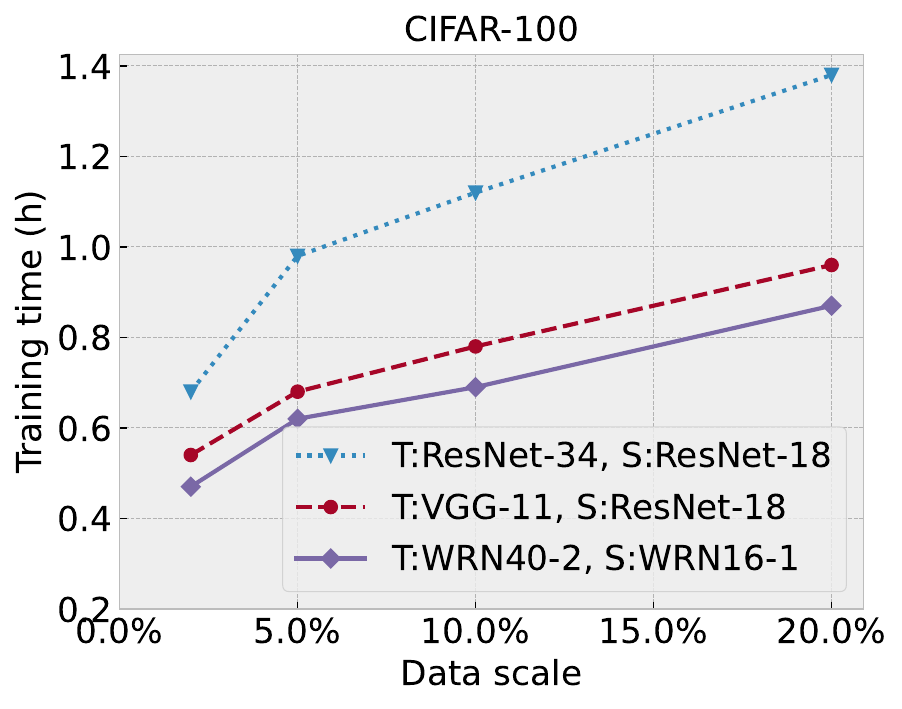}
    \end{tabular}
    \centering
    \caption{Performance comparison of SSD-KD under different synthetic data scales against the original training dataset size, in terms of top-1 classification accuracy (\%) and overall training time cost (hour).}
    \label{fig:scale}
\vspace{-0.5em}
\end{figure*}

\begin{table*}
\setlength{\abovecaptionskip}{0.1in}
\centering
\caption{Effect of two core modules in SSD-KD. PS: the priority sampling function, Difficulty$|$Diversity: the difficulty$|$diversity-aware balancing term in the modulating function, see Eq.~(\ref{eq:phi}).}
\resizebox{0.55\linewidth}{!}{%
\begin{tabular}{ccccccc}
  \toprule[1.2pt]
  Dataset                    & Baseline & PS & Difficulty & Diversity & \multicolumn{1}{c}{\begin{tabular}{@{}c@{}}ResNet-34 \\ResNet-18\end{tabular}} & \multicolumn{1}{c}{\begin{tabular}{@{}c@{}}WRN40-2 \\WRN16-1\end{tabular}}  \\
  \midrule
  \multirow{6}{*}{CIFAR-10}  & $\checkmark$ &     &  &   & 87.35(1.19h)   & 80.47(0.84h) \\
                             & $\checkmark$ & $\checkmark$ &  &   & 90.28(1.01h)  & 87.19(0.69h) \\
                             & $\checkmark$ &     & $\checkmark$& & 88.57(1.20h) & 83.12(0.86h) \\
                             & $\checkmark$ &     & & $\checkmark$& 89.12(1.21h) & 84.31(0.87h) \\
                             & $\checkmark$ &     & $\checkmark$& $\checkmark$& 89.95(1.21h) & 85.79(0.88h) \\
                             & $\checkmark$ & $\checkmark$ & $\checkmark$ & $\checkmark$ & 92.92(1.11h)  & 88.07(0.77h) \\
  \midrule
  \multirow{6}{*}{CIFAR-100} & $\checkmark$ &     &  &   & 64.05(1.19h) & 41.67(0.84h) \\
                             & $\checkmark$ & $\checkmark$ &  &   & 71.70(1.01h)  & 49.70(0.69h) \\
                             & $\checkmark$ &     & $\checkmark$& & 66.94(1.20h) & 43.96(0.86h) \\
                             & $\checkmark$ &     & & $\checkmark$& 68.52(1.21h) & 46.34(0.87h) \\
                             & $\checkmark$ &  &  $\checkmark$ & $\checkmark$ & 70.85(1.21h) & 47.82(0.88h) \\
                             & $\checkmark$ & $\checkmark$ & $\checkmark$ & $\checkmark$ & 73.05(1.12h) & 51.96(0.77h) \\
  \toprule[1.2pt]
  \label{tab:ablation}
\vspace{-1.5em}
\end{tabular}
}
\end{table*}

\begin{figure}
\setlength{\abovecaptionskip}{-1.pt}
    \centering
    \includegraphics[width=0.85\linewidth]{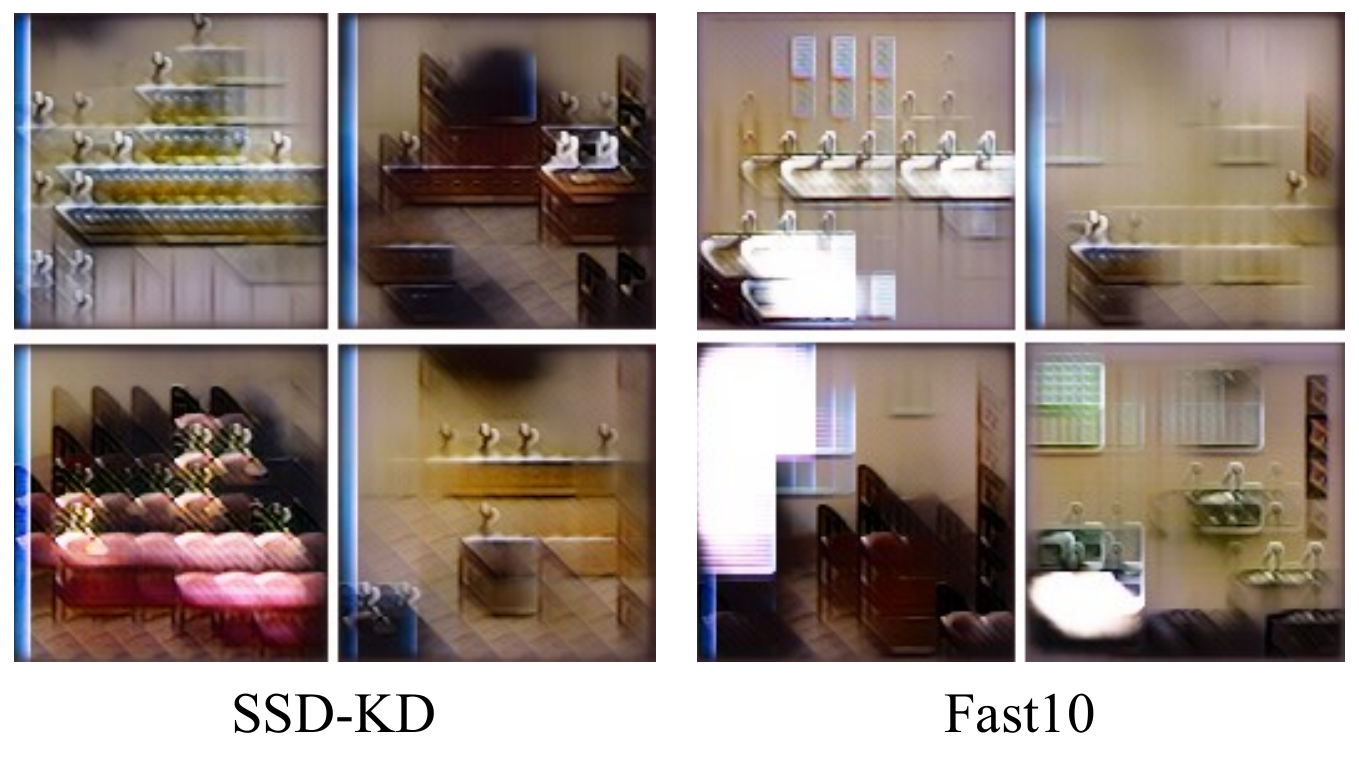}
    \caption{Visualization examples of synthetic image samples generated by Fast10 and our SSD-KD for the NYUv2 dataset.}
    \label{fig:vis001}
    \vspace{-1.5em}
\end{figure}

\textbf{Effect of the synthetic data scale.} So far, we have already demonstrated that
SSD-KD has the appealing capability to attain efficient and effective end-to-end distillation training using a small amount of synthetic samples. To better explore the boundary of this capability, we conduct an ablation with three teacher-student model pairs, including ResNet34$\rightarrow$ResNet18, VGG11$\rightarrow$ResNet18, and WRN40-2$\rightarrow$WRN16-1. 
In the experiments, we decrease the synthetic data scale used in SSD-KD~\cite{fasterfang} from 50,000(100\% relative to the original training data size) to \{10000(20\%), 5000(10\%), 2500(\%5), 500(1\%)\} for the CIFAR-10 and CIFAR-100 datasets. As shown in Fig.~\ref{fig:scale}, the accuracy of the student model trained by SSD-KD remains stable across a relatively large synthetic data scale range, for all teacher-student model pairs. When decreasing the synthetic data scale, the overall training time cost appears to decrease nearly linearly, while the student model accuracy drop is mild (less than 10\% even for the synthetic data scale 500).


\textbf{Effect of the core modules.} In Table \ref{tab:ablation}, we provide an ablation to scrutinize our SSD-KD systematically. We observe that: (1) the two modulating functions (consisting of a diversity-aware term and a difficulty-aware term, see Eq.~(\ref{eq:phi})) and the priority sampling function, are both critical to our SSD-KD; (2) the combination of them strikes a good tradeoff between model accuracy and training efficiency.


\textbf{Visualization of data inversion.} Fig. \ref{fig:vis001} shows examples of synthetic images inverted by Fast10 and our SSD-KD for the NYUv2 dataset. Compared to Fast10, our method can better invert texture information and has less noise.

\section{Conclusion}
\vspace{-0.5em}

In this paper, we presented SSD-KD, the first fully efficient method to advance adversarial data-free knowledge distillation research. Benefiting from a small-scale data inversion and sampling mechanism based on a modulating function and a priority sampling function, SSD-KD can flexibly balance class distributions in terms of synthetic sample diversity and difficulty during both data inversion and distillation processes, attaining efficient and effective data-free knowledge distillation. Extensive experiments on image classification and semantic segmentation benchmarks validate the efficacy of SSD-KD. We hope our work can inspire future research on efficient D-KD designs.

\section{Acknowledgement}
This work was supported by the National Natural Science Fund for Key International Collaboration (62120106005).

\clearpage

{
    \small
    \bibliographystyle{ieeenat_fullname}
    \bibliography{main}
}


\end{document}